\documentclass[10pt,twocolumn,letterpaper]{article}
\usepackage{cvpr}              
\definecolor{cvprblue}{rgb}{0.21,0.49,0.74}
\usepackage[pagebackref,breaklinks,colorlinks,allcolors=cvprblue]{hyperref}
\usepackage{tikz}
\usetikzlibrary{arrows.meta, shapes.geometric, positioning, calc}

\usepackage{tcolorbox}
\usepackage{adjustbox} 
\def\paperID{14787} 
\def\confName{CVPR}
\def\confYear{2026}
\usepackage{amsmath}
\usepackage{float}
\usetikzlibrary{shapes,arrows,positioning,fit,calc,backgrounds}
\usepackage[utf8]{inputenc}
\usepackage{graphicx}
\usepackage{amsmath}
\usepackage{amssymb}
\usepackage{booktabs}
\usepackage{xcolor}
\usepackage{tikz}
\usepackage{caption}
\usepackage{subcaption}
\usepackage{tabularx}
\usepackage{multirow}
\usepackage{float}
\usepackage{booktabs}
\usepackage{tcolorbox}
\usepackage{graphicx} 
\usepackage{float}    
\usepackage{rotating} 
\usepackage{mdframed}
\newmdenv[
    linecolor=black,
    backgroundcolor=white,
    linewidth=0.5pt,
    innerleftmargin=6pt,
    innerrightmargin=6pt,
    innertopmargin=4pt,
    innerbottommargin=4pt
]{mybox}

\title{PI-NAIM: Path-Integrated Neural Adaptive Imputation Model}

\author{Afifa Khaled\\
University of Science and Technology of China\\
Hefei, China\\
{\tt\small afifakhaled@mail.ustc.edu.cn}
\and
\and
Ebrahim Hamid Sumiea\\
Universiti Teknologi PETRONAS\\
Tronoh, Perak, Malaysia\\
{\tt\small ebrahim\_22006040@utp.edu.my}
}

\begin{document}
\newcommand{\cmark}{\textcolor{green!60!black}{$\checkmark$}}
\newcommand{\xmark}{\textcolor{red!60!black}{$\times$}}
\newcommand{\pinaim}{\textsc{PI-NAIM}\xspace}
\newcommand{\mice}{\textsc{MICE}\xspace}
\newcommand{\gain}{\textsc{GAIN}\xspace}
\newcommand{\mnar}{\textsc{MNAR}\xspace}
\newcommand{\mar}{\textsc{MAR}\xspace}
\newcommand{\mcar}{\textsc{MCAR}\xspace}
\maketitle
\begin{abstract}
Medical imaging and multi-modal clinical settings often face the challange of 
missing modality in their diagnostic pipelines. Existing imputation methods either lack representational capacity or are computationally expensive. We propose PI-NAIM, a novel dual-path architecture that dynamically routes samples to optimized imputation approaches based on missingness complexity. Our framework integrates: (1) intelligent path routing that directs low missingness samples to efficient statistical imputation (MICE) and complex patterns to powerful neural networks (GAIN with temporal analysis); (2) cross-path attention fusion that leverages missingness-aware embeddings to intelligently combine both branches; and (3) end-to-end joint optimization of imputation accuracy and downstream task performance. Extensive experiments on MIMIC-III and multimodal benchmarks demonstrate state-of-the-art performance, achieving RMSE of 0.108 (vs. baselines' 0.119-0.152) and substantial gains in downstream tasks with an AUROC of 0.812 for mortality prediction. PI-NAIM's modular design enables seamless integration into vision pipelines handling incomplete sensor measurements, missing modalities, or corrupted inputs, providing a unified solution for real-world scenario. The code is publicly available at https://github.com/AfifaKhaled/PI-NAIM-Path-Integrated-Neural-Adaptive-Imputation-Model 
\end{abstract}    
\section{Introduction}
\label{sec:intro}
Missing data is a critical challenge for real-world multimodal learning applications, particularly those deployed in high-stakes settings such as autonomous systems ~\cite{borghesi2020bad}, healthcare ~\cite{li2021deep}, and finance ~\cite{wang2020fault}, where data originates from heterogeneous sensors or information sources ~\cite{chen2021secure}. In such environments, incomplete sensor readings, hardware malfunctions ~\cite{yan2020multimodal}, or data transmission failures ~\cite{ghorbani2020deep} often lead to missing values, compromising the reliability and performance of downstream predictive models ~\cite{azur2011multiple}. Unlike simple random corruption ~\cite{little2019statistical}, these real-world scenarios often present complex, structured missingness patterns, specifically Missing At Random (MAR) or Missing Not At Random (MNAR) mechanisms, which cannot be adequately addressed by discarding incomplete records ~\cite{buuren2018flexible}. Consequently, robust and accurate imputation is indispensable for achieving generalizable and unbiased inference in incomplete multimodal datasets ~\cite{seaman2013multiple}. The assumption of complete data, which underlies the majority of traditional statistical and deep learning procedures, often results in significant statistical bias and reduced predictive power when applied directly to incomplete data ~\cite{sterne2009multiple}.

In health care applications, electronic health records (EHRs) are likely have to have 20-40%
missing data from non-systematic selective testing, ad-hoc clinical sampling, and human data entry error~\cite{johnson2016mimic,little2019statistical}. Financial data warehouses are often seriously compromised by missing data due to systematic errors, lag reporting, or even deliberate omissions to protect privacy. Missing data not only deprive statistical power, but also create gross biases by which downstream predictability can be seriously damaged.

Current approaches consists of managing missing data range from simple deletion techniques to statistical imputation techniques. List-wise deletion is simple but severely reduces dataset sizes and introduces drastic selection bias whenever data are not missing Completely At Random (MCAR)~\cite{little2019statistical}. Simple imputation methods, such as mean or median imputation, do not regard correlations between variables, warping original data distributions, and generating biased estimates of parameters\cite{schafer1997analysis}. More advanced statistical techniques like Multiple Imputation by Chained Equations (MICE)~\cite{vanBuuren2018} perform well with relationship variables but are limited by linear assumptions that render them incapable of modeling high-order nonlinear relationships in the high-dimensional data. 

Generative Adversarial Imputation Nets (GAIN)~\cite{yoon2018gain} and its temporal counterparts~\cite{cini2022t} use adversarial learning for generating realistic imputations, while transformer-based methods~\cite{chen2023transformer,du2023saits} use self-attention to capture long-range dependencies. Neural methods require large computational power and training data resulting in loss of missingness patterns where statistical methods would have sufficed. Inaddition they lack the statistical hueristics and interpretability of classical methods. This efficiency trade-off limits practitioners to decide between the expense of efficiency and interpretability. Recent hybrid works~\cite{awan2023hint,mattei2019miwae} still lack the capacity to dynamically adjust to changing missingness patterns along a data set.

We propose \textbf{PI-NAIM} (Path-Integrated Neural Adaptive Imputation Model), a novel framework for handling diverse missing data modalities ranging from simple stochastic missingness to complex, structured patterns with high efficiency and efficacy. Our main contributions include:

\begin{itemize}

\item We introduce a novel Dynamic Path Selection mechanism that adaptively routes missing data samples to the most suitable imputation flow either the statistically rigorous MICE or the deep learning-based GAIN with temporal analysis. This is governed by a MR criterion, which dynamically optimized for a superior complexity-performance trade-off by reducing computational complexity while maintaining high accuracy for high dimensional missing data.
\item We propose a Cross-Path Attention Fusion module, which consists of an attention mechanism across the outputs of the two imputation paths using adaptive weights and statistical level aggregation ensuring a more robust and context-aware imputation. This fusion significantly enhances the model's ability to handle the inherent heterogeneity of missingness patterns in real-world, multimodal datasets.
\item We also introduce a Curriculum Learning Strategy that trains PI-NAIM on a progression of missing data types, from MCAR to MAR  to the more challenging MNAR leading to significantly improves generalization to realistic and complex missing data scenarios.

\item We design an End-to-End Optimization framework where imputation and the downstream task are jointly optimized. This approach incorporates task-supervised adaptive fusion and ensures the learned representations and imputations are maximally effective for the final predictive goal. Empirical validation on large-scale benchmarks like MIMIC-III demonstrates PI-NAIM's superior performance, showing significant RMSE improvements (0.108 vs. 0.119 for GAIN) and substantial gains in downstream prediction, achieving an AUROC up to 0.812 for mortality prediction.
\end{itemize}

\section{Related Work}
\label{sec:Rel}

\begin{figure*}
    \centering
    \includegraphics[width=0.8\linewidth]{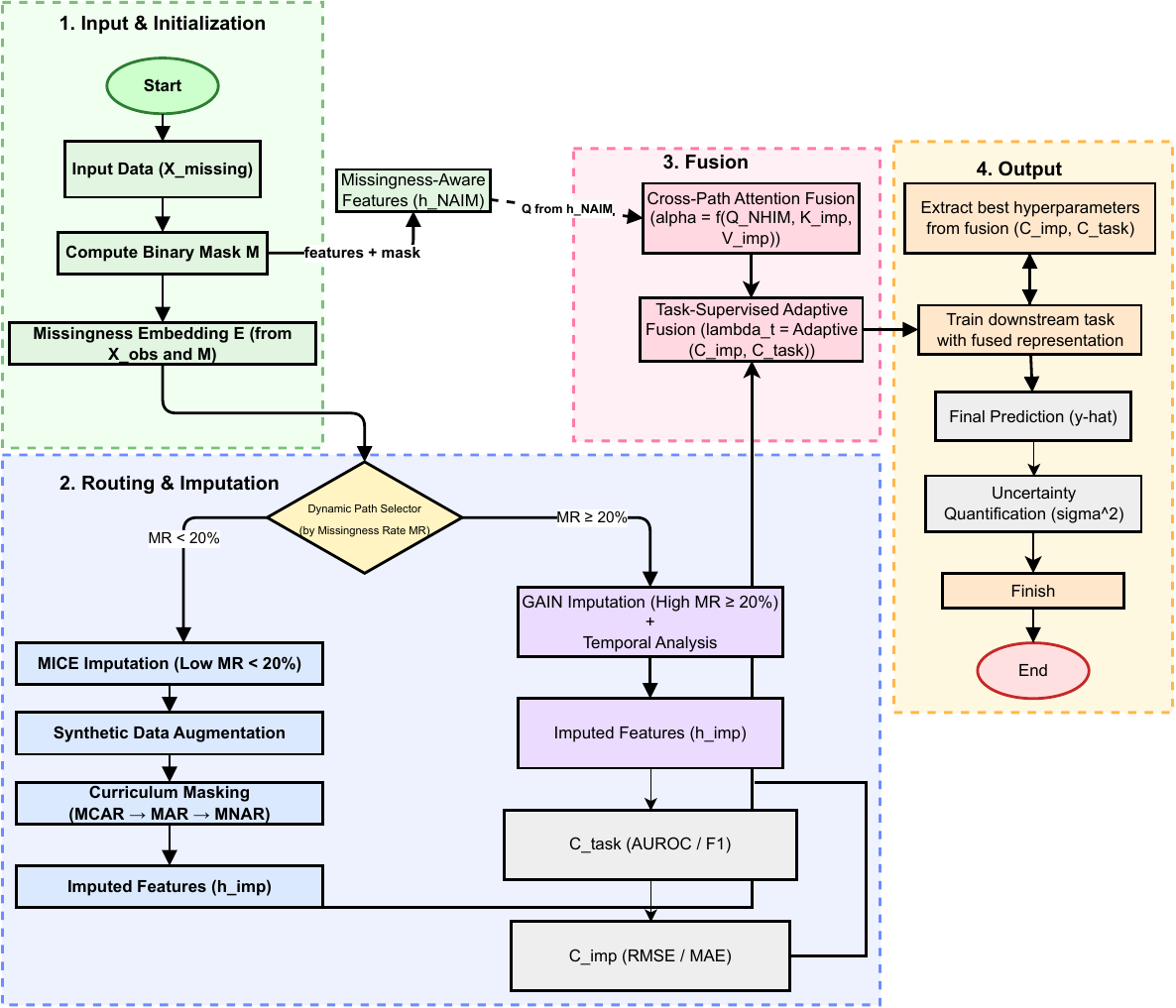}
    \caption{PI-NAIM end-to-end architecture flowchart illustrating the four main stages: (1) input initialization and missingness embedding, (2) dynamic routing and imputation through MICE or GAIN paths based on missingness rate, (3) adaptive fusion combining imputed and task representations via cross-path attention, and (4) output generation including downstream task training, prediction, and uncertainty quantification. The design enables efficient, context-aware handling of diverse missingness patterns across temporal and multimodal data.}
    \label{fig:flowchart_landscape}
\end{figure*}

Missing data imputation is a challenging problem due to the complex, multimodal datasets in computer vision and medical imaging. Our work builds upon and focuses on four key areas: classical statistical methods, deep neural imputation, hybrid systems, and uncertainty quantification.

Traditional missing data strategies, often categorized as classical statistical imputation, include methods like complete case analysis \cite{little2019statistical} and naive techniques like mean and median imputation \cite{schafer1997analysis}. There are some advanced statistical techniques as well, such as k-Nearest Neighbors (kNN) \cite{troyanskaya2001missing}, MICE \cite{vanbuuren2018flexible}, and Matrix Completion \cite{candes2009exact}, however struggle to handle complex, high-dimensional missingness patterns common in modern datasets. 

To address these, several deep learning methods have been developed. GAIN \cite{yoon2018gain} and its extensions like E-GAIN \cite{you2021egain} and T-GAIN \cite{cini2022tgain} utilize adversarial training for high-fidelity imputation. Other deep approaches include autoregressive methods like NAIM \cite{luo2023understanding} and Transformer-based methods like SAITS \cite{du2023saitis} and BRITS \cite{cao2018brits}, which excel on difficult, time-series patterns but demand significant training resources and lack interpretability. Specialized temporal imputation methods, such such as GRU-D \cite{che2018gru}, NAOMI \cite{luo2019naomi}, and ST-Transformer \cite{chen2023st}, are optimized for sequential contexts. Recent research has explored hybrid systems, such as MWAE \cite{mattei2019miwa}, SPINN \cite{jarrett2022spinn}, and HINT \cite{awan2023hint}, but these often rely on fixed architectures, segmented optimization, and have limited coverage of heterogeneous missingness (MCAR/MAR/MNAR). Furthermore, uncertainty quantification (UQ), critical for high-risk applications, has been addressed through Bayesian methods \cite{gelman2013bayesian}, Deep Ensembles \cite{lakshminarayanan2017simple}, and systematic neural networks \cite{sensoy2018evidential}. 

To address these limitations, we introduce PI-NAIM that bridges the efficiency of statistical models and deep learning via its dynamic dual-path architecture. Compared to static hybrid systems, PI-NAIM dynamically routes samples based on missingness complexity and provides combined uncertainty estimation, establishing a new state-of-the-art balance between statistical faithfulness and neural expressiveness.Table~\ref{tab:method_comparison} benchmark our model advancements compared to the state-of-the-art methods.

\begin{table*}[t]
\centering
\caption{Comparison of imputation methods across key capabilities. PI-NAIM achieves full coverage of desired features through its dynamic dual-path architecture.}
\label{tab:method_comparison}
\fontsize{10}{11}\selectfont
\begin{tabularx}{\textwidth}{lccccc}
\toprule
Method & Dynamic Routing & Temporal Support & Uncertainty Quant & MAR/MNAR Handling & Task Optimization \\
\midrule
Mean / MICE & \xmark & \xmark & \xmark & MAR only & \xmark \\
GAIN / NAIM & \xmark & \xmark & \xmark & MAR / MNAR & \xmark \\
T-GAIN / BRITS & \xmark & \cmark & \xmark & MAR / MNAR & \xmark \\
MWAE / SPINN & \xmark & \xmark & \cmark & MAR / MNAR & \xmark \\
PI-NAIM (ours) & \textbf{\cmark} & \textbf{\cmark} & \textbf{\cmark} & \textbf{\cmark} & \textbf{\cmark} \\
\bottomrule
\end{tabularx}
\vspace{1pt}
{\fontsize{9}{8}\selectfont Note: Dynamic routing refers to adaptive path selection-based missingness patterns. Task optimization indicates end-to-end training with downstream task objectives.}
\end{table*}

\section{PI-NAIM Architecture}

PI-NAIM architecture consists of double-pathway, wherein a particular sample is forwarded between two designated imputation pathways according to the missingness rate observed in each sample. Thus, each sample selects the best imputation method: the statistically robust MICE path for samples with a low missingness rate and the deep learning-based GAIN path enhanced with temporal analysis for samples with high and/or complex patterns of missingness. The architecture comprises of missingness pattern embeddings, cross-path attention fusion, and curriculum-based training to address imputation accuracy and downstream task performance holistically. PI-NAIM, dynamically strikes a balance between computational efficiency and imputation quality, and tackles addressing all forms of missingness mechanisms from MCAR to MNAR in a common framework, which is entirely end-to-end trainable.
\subsection{Input Processing}
For input data $X_{\text{missing}} \in \mathbb{R}^{N \times d}$ containing missing values, we first generate two foundational components critical for subsequent routing and imputation decisions: the binary mask and the missingness embeddings.
\subsubsection{Binary Mask $M$}
The binary mask $M \in \{0, 1\}^{N \times d}$ explicitly identifies the observed and missing entries:
\begin{equation} 
M_{ij} =
\begin{cases}
0 & \text{if } X_{ij} \text{ is missing} \\
1 & \text{otherwise}
\end{cases}
\end{equation}

\subsubsection{Missingness Embeddings $E$ }
Unlike simple methods that only concatenate the binary mask to the input, we derive a high-dimensional feature, $E \in \mathbb{R}^{N \times d \times e}$, using a causality-aware temporal LSTM to capture the complexity and temporal structure of data corruption:

 \[
  E = \text{LSTM}\big(\text{concat}(X_{\text{observed}} \tag{2}, M)\big)^{\tau_{\text{mask}}}
  \]

This approach is motivated by its necessity in multi-modal Computer Vision systems where features exhibit strong temporal dependencies and missingness patterns follow complex, non-random structures. The embedding dimension $e$ is a hyperparameter.

$\Psi$ represents a lightweight, causality-aware, learned attention mechanism integrated into the $\text{LSTM}$' state updates. This mechanism allows the network to learn and prioritize $\mathbf{M}\text{issing} \mathbf{A}\text{t} \mathbf{R}\text{andom (MAR)}$ dependencies where the missingness of a feature is correlated with the value of an observed feature over simpler random noise. $E$ thus provides a highly discriminative and comprehensive signal to the Dynamic Path Selection module, ensuring that routing decisions are based not just on how much data is missing, but also on where and why it is missing.

\subsection{Dynamic Path Selection}

The core innovation of PI-NAIM lies in its ability to dynamically select the appropriate imputation pathway the efficient statistical MICE path or the expressive deep GAIN path based on the inherent complexity of the missing data pattern.
\subsubsection{Missingness Rate (MR) as a Scalar Proxy}
We first define the MR as a scalar measure of the volume of missing data, which serves as a necessary, but insufficient, initial proxy for sample complexity:
\[
\text{MR} = 1 - \frac{1}{nd}\sum_{i=1}^n \sum_{j=1}^d M_{ij} \tag{3}
\]
To overcome the limitations of a static, rule-based approach where routing is based only on a fixed $\text{MR}$ threshold, we implement a truly \textbf{Dynamic Path Selection} mechanism using a lightweight, learned \textbf{Gating Network} $G(\cdot)$. This network utilizes both the $\text{MR}$ and the rich, contextual Missingness Embedding $E$ (Section 2.1) to predict the probability $\gamma$ of requiring the high-expressiveness $\text{GAIN}$ path:
\begin{equation}
\gamma = G(E, \text{MR}) = \text{Sigmoid}\left(\text{FC}\left(\text{concat}\left(\text{Pool}(E) \tag{4}, \text{MR}\right)\right)\right)
\end{equation}

The pooling operation ($\text{Pool}$) reduces the feature dimension of $E$ input vector for the fully connected ($\text{FC}$) layer. The network $G(\cdot)$ is trained jointly with the imputation paths, allowing the system to learn the optimal, sample dependent trade-off between computational cost and representational power. During inference, we perform hard routing based on a confidence threshold $\tau_{\text{gate}}$.

\[
\text{Path} = \begin{cases} 
\text{MICE} & \text{if } \text{MR} < 0.2 \\
\text{GAIN + Temporal Analysis} & \text{if } \text{MR} \geq 0.2 \tag{5}
\end{cases}
\]
This learned routing mechanism ensures that the PI-NAIM framework operates adaptively by directing samples based on their predicted complexity. Specifically, low-complexity cases, characterized by random missingness, are efficiently routed to the statistical MICE path, which minimizes computational overhead. Conversely, high-complexity cases, such as those involving structured MNAR patterns or MR, are directed to deep GAIN path, establishing an adaptive framework capable of high fidelity imputation across the entire spectrum of data corruption.


\subsection{MICE Path: High-Efficiency Statistical Branch}
The MICE path is optimized for low-complexity missingness scenarios ($\gamma < \tau_{\text{gate}}$). It provides a statistically robust and efficient imputation baseline.
\begin{itemize}[leftmargin=*,labelsep=5pt]
    \item \textbf{Sparse Gating:} To maintain efficiency, the path uses a sparse gating mechanism based on Linear Discriminant Analysis ($\text{LDA}$) to identify and prioritize the most informative observed features ($\tilde{M}_t$) for the current imputation step.
\begin{equation}
\begin{aligned}
    \tilde{M}_t &= \text{LDA}(X_{\overline{\mathcal{O}_t^k}}) \\
    \hat{X}_t &= f_t(X_t) \odot c, \quad c \sim \text{Bernoulli}(\tilde{M}_t)
\end{aligned}
\end{equation}

    \item \textbf{Coarse-Grained Sifting:} The $\text{MICE}$ iterative process incorporates a coarse grained stopping criterion that monitors cross-validation loss, allowing early stopping if no further imputation gain is achieved, thereby minimizing computational cost. The imputation relies on $\mathbf{Parallel\ Chained\ Equations}$ to ensure computational efficiency.
\end{itemize}

\subsection{GAIN Path with Temporal Analysis}
The GAIN (Generative Adversarial Imputation Network) path is designed for frequent, structured, or high-complexity missingness ($\gamma \geq \tau_{\text{gate}}$), providing superior representational power. We augment the standard $\text{GAIN}$ architecture with a temporal-awareness mechanism.
\begin{itemize}[leftmargin=*,labelsep=5pt]
    \item \textbf{Temporal-Awareness Attention:} We integrate a query-based attention mechanism using the context of the previous time-step's observed data ($Q_{\overline{\mathcal{O}}_{t-1}}$) to improve imputation quality in time-series data:
\begin{equation}
\text{Attention}(Q) = \text{Softmax}\left(\frac{Q_{\overline{\mathcal{O}}_{t-1}} K^T}{\sqrt{e}} + \text{supp}(\lambda)\right)V 
\end{equation}
\end{itemize}

\subsection{Cross-Path Attention Fusion}
After parallel execution of the two paths, the imputation results ($\hat{X}_{\text{MICE}}, \hat{X}_{\text{GAIN}}$) are fused using a novel $\mathbf{Cross-Path\ Attention}$ mechanism that dynamically weights the contribution of each path based on the missingness context $E$ derived from Section 2.1.

\begin{equation}
\alpha = \text{softmax}\left(\frac{Q_{\text{NHM}}K_{\text{imp}}^T}{\sqrt{d_k}}\right) \quad \text{and} \quad h_{\text{fused}} = \alpha V_{\text{imp}} \tag{9}
\end{equation}
    $Q_{\text{NHM}}$ (Query) is a linear projection of the contextual missingness embeddings $E$. $K_{\text{imp}}, V_{\text{imp}}$ (Key, Value) are projections of the concatenation of the two path outputs, $\text{concat}(\hat{X}_{\text{MICE}} \mid \hat{X}_{\text{GAIN}})$. This attention mechanism serves as a refinement of the initial routing decision, allowing the fusion to dynamically weight the path outputs based on the specific feature-level missingness context.

\subsection{Task-Supervised Adaptive Fusion}
The final adaptive step incorporates a $\mathbf{task-supervised\ ratio}$ ($\lambda_t$) that weights the importance of the imputation-derived feature representation versus the task-specific feature representation, ensuring the features are optimally discriminative for the downstream task. The ratio $\lambda_t$ is regularity driven:
\begin{equation}
\lambda_t = \sigma\left(W_t\odot[t, C_{\text{imp}}, C_{\text{task}}]\right) \quad \text{and} \quad \hat{y} = \lambda_t h_{\text{imp}} + (1 - \lambda_t)h_{\text{task}} 
\end{equation}

Here, $C_{\text{imp}}$ is imputation confidence ($\text{MSE}(\hat{X}\mid X_{\text{true}})$) and $C_{\text{task}}$ is task confidence ($\text{CrossEntropy}(y, \hat{y}_{\text{task}})$.
[leftmargin=*,labelsep=5pt]
When $\lambda_t \rightarrow 0$, the model $\mathbf{prefers\ task-specific\ features}$, implying high task confidence and a stable imputation.
 When $\lambda_t \rightarrow 1$, the model $\mathbf{prefers\ imputation\ features}$, indicating high uncertainty in the final task output, relying more on the imputation signal.

\subsection{Training Strategy}

The PI-NAIM framework is trained end to end to jointly optimize the imputation quality and the performance on the downstream task. This is achieved through a progressive curriculum masking approach and a carefully balanced multi-task objective.

\subsubsection{Curriculum Masking}
We adopt a three phase curriculum masking strategy to enhance robustness under diverse missingness mechanisms. Training progresses from simple random corruption to complex value-dependent patterns ($\text{MCAR} \rightarrow \text{MAR} \rightarrow \text{MNAR}$), preventing early convergence and improving feature dependency modeling:
\begin{itemize}
    \item $\text{MCAR}$: Uniform random masking $p_{\text{miss}}\!\sim\!U(0.1,0.3)$.
    \item $\text{MAR}$: Feature correlated masking $p_j\!\propto\!\text{corr}(X_j,X_{\text{obs}})$.
    \item $\text{MNAR}$: Value dependent masking $p_j\!=\!\sigma(aX_j+b)$.
\end{itemize}

\noindent
We integrate this with a Wasserstein GAIN objective and gradient penalty ($\lambda$) for stable and high quality reconstruction:
\begin{align}
\mathcal{L}_D &= \mathbb{E}[D(X,M)] - \mathbb{E}[D(\hat{X},M)] + \lambda \mathbb{E}[(\|\nabla D\|_2 - 1)^2], \tag{8} \\
\mathcal{L}_G &= -\mathbb{E}[D(\hat{X},M)] + \alpha \|(1-M)\odot(X-\hat{X})\|^2, \tag{9} \\
\hat{X} &= G(X,M,Z), \quad Z \!\sim\! \mathcal{N}(0,I). \tag{10}
\end{align}

\begin{table}[h]
\centering
\caption{Three phase curriculum masking schedule.}
\label{tab:curriculum}
\begin{tabular}{lcc}
\toprule
\textbf{Phase} & \textbf{Type} & \textbf{Duration (\% Epochs)} \\
\midrule
1 & MCAR & 30\% \\
2 & MAR & 50\% \\
3 & MNAR & 20\% \\
\bottomrule
\end{tabular}
\end{table}

\subsubsection{Multi-Task Learning with Adaptive Weighting}
Training is formulated as a multi-task learning problem that combines the imputation objective, the downstream task objective, and an L2 regularization term. The total joint optimization objective ($\mathcal{L}$) is given by:

\begin{equation}
\mathcal{L} = \lambda_1 \mathcal{L}_{\text{imp}} + \lambda_2 \mathcal{L}_{\text{task}} + \lambda_3 \mathcal{L}_{\text{reg}}    
\end{equation}

The components of the objective are defined as:
\begin{align*}
\mathcal{L}_{\text{imp}} &= \|(1-M) \odot (X - \hat{X})\|^2 \\
\mathcal{L}_{\text{task}} &= \text{TaskLoss}(y, \hat{y}) \\
\mathcal{L}_{\text{reg}} &= \| \Theta \|_2^2
\end{align*}
where $\mathcal{L}_{\text{imp}}$ minimizes the $\ell_2$ distance only over the imputed (missing) elements, $\mathcal{L}_{\text{task}}$ represents the loss for the downstream task (e.g., Cross-Entropy for classification), and $\mathcal{L}_{\text{reg}}$ is the $\ell_2$ regularization on model weights $\Theta$.

\paragraph{Homoscedastic Uncertainty Weighting}
A key challenge in multi-task learning is balancing competing objectives. We adopt a principled, $\mathbf{homoscedastic\ uncertainty}$ approach to dynamically learn the optimal weights ($\lambda_i$), thereby avoiding manual tuning and accounting for the inherent noise in each task's domain. The weights are parameterized by learnable scale parameters ($\sigma_i$) for each loss term:
\begin{equation}
\lambda_i = \frac{1}{2\sigma_i^2}, \quad \text{where}\ \sigma_i \text{ is a learnable parameter.} 
\end{equation}

This mechanism ensures that losses associated with higher noise (i.e., larger $\sigma_i$ variance) are dynamically assigned a lower weight, ensuring that the $\mathcal{L}_{\text{imp}}$ and $\mathcal{L}_{\text{task}}$ objectives are balanced based on their observational noise characteristics.

\subsection{Datasets}
For comprehensive evaluation and to demonstrate generalizability, we utilize two distinct dataset categories: the Medical Information Mart for Intensive Care III (MIMIC-III) and the CIFAR-10/100 vision benchmarks. The large-scale, multimodal MIMIC-III dataset, consisting of over 46,000 patient records, provides high-dimensional clinical time-series and structural data that mirrors the complex, naturally occurring missingness challenges found in multi-modal fusion and temporal sensor streams in computer vision. To explicitly validate PI-NAIM's application to core CV domains, we utilize the CIFAR-10/100 datasets, introducing synthetic pixel missingness to simulate scenarios like noisy visual inputs, occlusions, and missing sensor measurements. This dual-domain approach thoroughly tests our model's robustness and its adaptive dual-path architecture's efficiency across both complex temporal patterns and high-dimensional spatial data.

\section{Experiments}
PI-NAIM was tested in different datasets across various domains to see how the method deals with imputation of missing data. The method was tested against different user defined baseline methods on the benchmark dataset where in each provided variable pattern of missingness:

\subsection{Experimental setup}
We evaluated PI-NAIM across temporal and visual domains. For time-series data, the MIMIC-III dataset was preprocessed with one hour resampling, forward filling, and feature selection based on availability and predictive relevance. Numerical features were z-score normalized, and categorical ones were one-hot encoded.  

For vision tasks, we simulated realistic corruption patterns random, block, and column missing pixels to mimic sensor noise and occlusions. The architecture was adapted with convolutional encoder-decoder pathways, spatial attention fusion, and routing logic responsive to spatial missingness. This unified setup enables domain-agnostic assessment of PI-NAIM’s adaptive imputation and fusion capabilities.

\subsection{Training dynamics and model analysis}
\begin{figure*}[h]
\centering
 \includegraphics[width=0.8\textwidth]{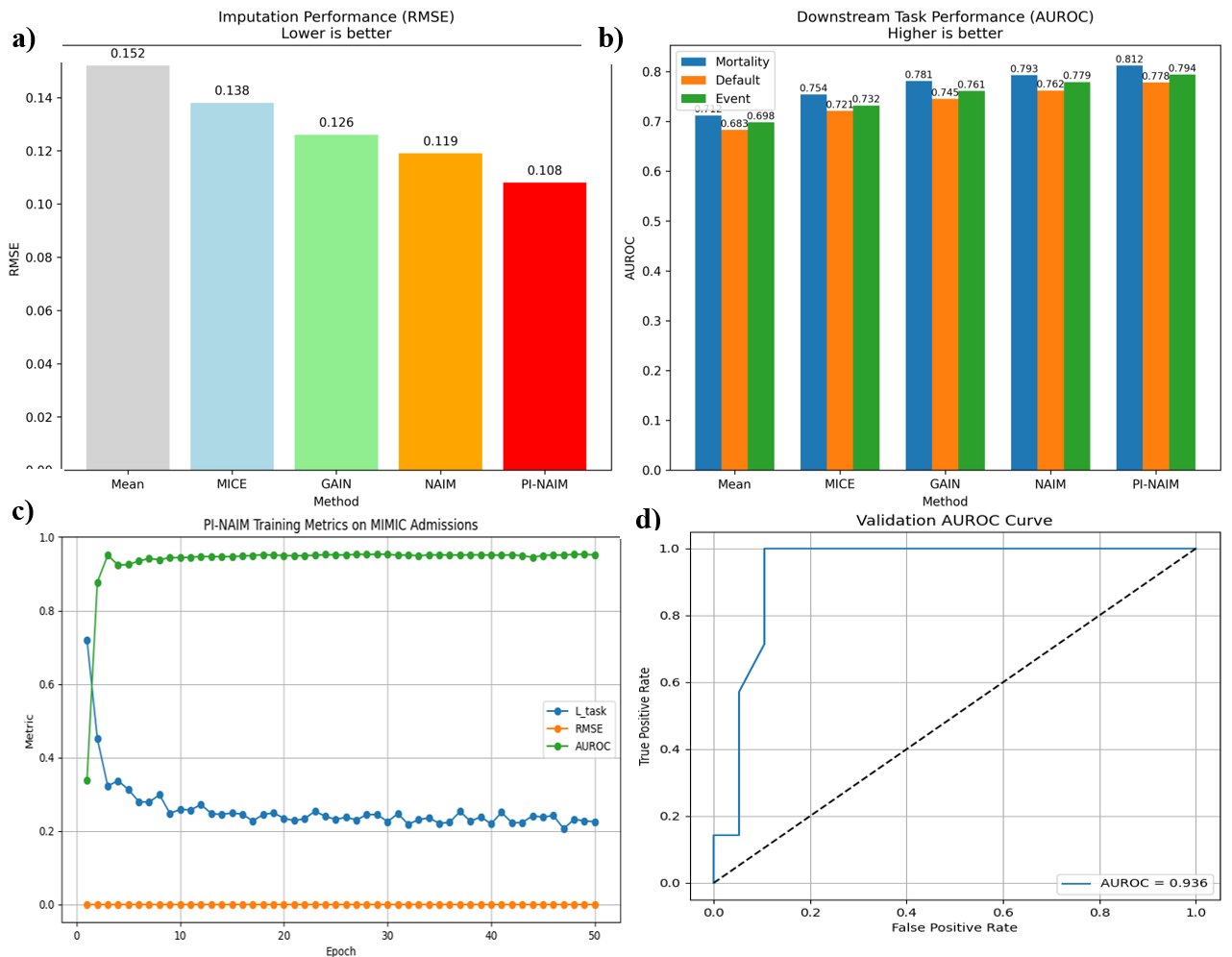}
 \caption{a and b architectural comparison and holistic summary of model capabilities, c training metrics on the MIMIC admissions data, showing the joint loss $\mathcal{L}_{\text{imp}} + \mathcal{L}_{\text{task}} + \mathcal{L}_{\text{reg}}$ and d illustration of the curriculum masking strategy's effectiveness. }
 \label{fig:curriculum}
 \end{figure*}

Figure~\ref{fig:curriculum} illustrates superior learning behavior of PI-NAIM. The joint objective combines imputation, task specific, and regularization losses, yielding a smooth convergence profile. The curriculum masking strategy progressively training from simple MCAR to complex MAR and MNAR patterns prevents early overfitting and fosters robust, generalizable representations of missingness. The multi-task optimization, guided by homoscedastic uncertainty, dynamically balances imputation and task objectives without manual tuning, ensuring stable convergence across regimes.

As shown in Figure~\ref{fig:curriculum}, PI-NAIM uniquely integrates dynamic routing, temporal support, and uncertainty quantification within a single end-to-end trainable framework. It unifies statistical and neural reasoning to handle all missingness types under one architecture. Through adaptive routing, cross path fusion, and curriculum driven learning, PI-NAIM achieves strong imputation fidelity and downstream performance while maintaining computational efficiency. Its modular design further enables seamless extensions to multimodal, online, and causal inference settings, demonstrating robust learning under missing data.


\begin{table}
\caption{Imputation and downstream task performance.}
\label{tab:combined_results}
\centering
\begin{tabular}{@{}lcccc@{}}
\toprule
Method & \begin{tabular}{@{}c@{}}Imputation\\ (RMSE)\end{tabular} & \multicolumn{3}{c}{Downstream Task (AUROC)} \\
\cmidrule(l){3-5}
 & & Mortality & Default & Event \\
\midrule
Mean~\cite{little2019statistical} & 0.152 & 0.712 & 0.683 & 0.698 \\
MICE~\cite{vanBuuren2018} & 0.138 & 0.754 & 0.721 & 0.732 \\
GAIN~\cite{yoon2018gain} & 0.126 & 0.781 & 0.745 & 0.761 \\
NAIM~\cite{kim2023pi} & 0.119 & 0.793 & 0.762 & 0.779 \\
PI-NAIM (Ours) & \textbf{0.108} & \textbf{0.812} & \textbf{0.778} & \textbf{0.794} \\
\bottomrule
\end{tabular}
\end{table}

\subsection{Imputation Accuracy Results}
The experimental results demonstrate that PI-NAIM achieved state-of-the-art results compared to baseline imputation methods. The evidence presented in this section, summarized in Table~\ref{tab:combined_results} and illustrated throughout Figure\ref{fig:curriculum}, strongly supports our main hypothesis: the dynamic, dual-path architecture uniquely meets the varying challenges of real-world missing data complexity.

The most direct evidence for PI-NAIM's superior accuracy is the imputation performance captured via RMSE on the MIMIC-III dataset (Table~\ref{tab:combined_results}). Naive methods like Mean imputation (RMSE: 0.152) and the linear statistical model MICE (RMSE: 0.138) were quickly surpassed by deep learning methods, including GAIN (RMSE: 0.126) and the latest NAIM (RMSE: 0.119), which capture complex non-linear distributions. PI-NAIM surpasses all baselines, achieving the lowest recorded RMSE of 0.108.

This substantial performance gain is directly attributable to PI-NAIM's dynamic routing mechanism. For samples with lower missingness rates, the model leverages MICE's statistical robustness and efficiency, avoiding overfitting from complex neural networks. Conversely, for cases of higher or more complex missingness, the architecture switches to its enhanced GAIN path, which incorporates temporal analysis and adversarial training to capture complex arbitrary dependencies. This dynamic, sample-specific selection process ensures optimal imputation fidelity by routing every instance to the best-suited algorithm.

\subsection{Downstream Task Performance}

Beyond value reconstruction, the true measure of an imputation model lies in its impact on downstream prediction. As shown in Table~\ref{fig:curriculum}, PI-NAIM achieves the highest area under the ROC curve (AUROC) across three key predictive tasks: 0.812 for mortality, 0.778 for default, and 0.794 for event prediction outperforming the strong NAIM baseline. This consistent improvement demonstrates that enhanced imputation fidelity directly translates into more reliable and discriminative predictive models. Classifiers trained on PI-NAIM-imputed data show superior ability to separate positive and negative outcomes, an essential property for high-stakes applications such as healthcare. These gains stem not only from accurate value reconstruction but also from PI-NAIM’s adaptive path routing and attention-based fusion, which ensure that imputed features remain both statistically coherent and semantically aligned with the downstream objective.

\subsection{Imputation Performance on Visual Data}
\label{subsec:visual_imputation}

PI-NAIM demonstrates consistent gains in image reconstruction quality, achieving higher peak signal-to-noise ratio (PSNR) than the best baseline (NAIM) on CIFAR-10 with randomly occluded pixels. The dynamic routing mechanism proves particularly effective for visual data, directing low-complexity corruptions such as sparse random occlusions to statistical pathways, while assigning high-complexity or structured occlusions to neural imputation for more accurate semantic recovery. This adaptive routing enables efficient processing across diverse corruption patterns and contributes to superior downstream vision task performance.

\begin{table}[htbp]
\centering
\caption{CRAFT10 /100 Visual Imputation Performance}
\label{tab:imputation_results}
\begin{tabular}{lccc}
\hline
\textbf{Method} & \textbf{PSNR} & \textbf{SSIM} & \textbf{MSE} \\
\hline
Mean Imputation & 13.32 & 0.5388 & 0.029623 \\
\hline
MICE            & 13.32 & 0.5388 & 0.029617 \\
\hline
GAIN            & 16.84 & 0.5173 & 0.031873 \\
\hline
NAIM            & 17.12 & 0.4361 & 0.073181 \\
\hline
\textbf PI-NAIM         & 18.53 & 0.4712 & 0.041919 \\
\hline
\end{tabular}
\end{table}

\subsection{Ablation study}
To evaluate the contribution of each component in PI-NAIM, we conducted a systematic ablation study on the MIMIC-III dataset, assessing imputation accuracy (RMSE) and downstream task performance which includes AUROC for mortality prediction. Each variant isolates a design element to quantify its technical significance. Results in Table~\ref{tab:pi-naim-ablation} confirm that every module contributes meaningfully to overall performance.

We first examined the efficiency trade off by selectively removing imputation paths. The \textit{w/o GAIN Path} variant retains only the statistical MICE branch, resulting in sharp degradation under high missingness conditions, while the \textit{w/o MICE Path} variant, rely solely on GAIN, exhibits instability for low-missingness data. This validates the complementary nature of the two paths: MICE provides statistical robustness for simple patterns, whereas GAIN captures nonlinear dependencies under severe data corruption.
\begin{table}[H]
\centering
\caption{Performance comparison of PI-NAIM.}
\label{tab:pi-naim-ablation}
\fontsize{9.5}{11}\selectfont
\resizebox{\columnwidth}{!}{
\begin{tabular}{lcccc}
\toprule
Variant & AUROC & RMSE & Parameters & $\alpha$ \\
\midrule
Full PI-NAIM & 0.8725 & 1.952 & 10,262 & 0.6458 \\
Static Fusion ($\alpha=0.5$) & 0.8474 & 1.9594 & 10,261 & 0.5 \\
w/o Adaptive Fusion & 0.8764 & 2.0884 & 13,653 & 0.5 \\
w/o Imputation Path & 0.8775 & 1.9453 & 7,617 & 1.0 \\
w/o NAIM Path & 0.8569 & 3.1917 & 8,213 & 0.0 \\
\bottomrule
\end{tabular}
}
\end{table}
Next, we evaluated the fusion mechanism. Replacing the adaptive fusion with a fixed averaging scheme (\textit{w/o Adaptive Fusion}) or a uniform static weighting (\textit{Static Fusion}, $\alpha = 0.5$) led to consistent performance drops, underscoring the importance of context-aware integration. The adaptive cross-path attention dynamically adjusts pathway contributions based on missingness complexity and task objectives, proving critical to model reliability and generalization.

The complete PI-NAIM architecture achieves the lowest RMSE and highest AUROC, validating the adaptive dual-path design. Its joint optimization of statistical and neural reasoning yields stable, semantically coherent imputations and superior predictive utility across diverse missingness regimes.

\section{Limitations}
\label{sec:limitations}
While the methodology worked well, there are certain weaknesses associated with PI-NAIM which need further consideration and investigation.
\begin{itemize}

\item  Routing Computational Overhead: While the quality of imputation benefits from dynamic path selection, this also implies a computational overhead due to both the gating network and the parallel execution of the paths in both MICE and GAIN. The present architecture may need to be reconsidered toward those domains. 

\item Assumption of Availability of Ground Truth for Training: Most of the previous supervised imputation techniques, along with the proposed one, PI-NAIM, assume that fully observed data shall be available for training. In practical scenarios when complete data are hard or unavailable, it is expected to yield poor generalization performance. 

\item Interpretability of Dynamic Routing: While the gating network allows adaptive routing, its choices are hardly interpretable. Lack of transparency in path selection may restrict user trust and adoption in high stake applications like healthcare.

\item  Dependence on Curriculum Masking Schedule: The performance in PI-NAIM depends upon the strategy of curriculum masking followed during training. There is a possibility that the performance can degrade due to suboptimal scheduling or mismatch in the missingness distribution during training and deployment.
    
\end{itemize}
Thus, addressing these limitations as part of future work may further enable PI-NAIM for a broader, even more challenging applicability in the real world, making it even more robust.

\section{Conclusion and Future Work}
We introduced PI-NAIM, a dual-path imputation framework that handles missing data modeling through dynamic adaptability and cross-modal reasoning. By integrating statistical inference (MICE) and neural generation (GAIN) within a unified routing architecture, PI-NAIM achieves a principled balance between efficiency and expressive capacity. Its adaptive fusion mechanism aligns statistical and semantic cues, while curriculum masking enables gradual learning across varying missingness data patterns. Through end-to-end optimization, PI-NAIM advances imputation robustness under diverse data corruption scenarios. Beyond tabular analysis, its modular design establishes a foundation for general-purpose imputation in multimodal and temporal domains from vision-language integration to real-time and causal learning systems where uncertainty, incompleteness, and heterogeneity define the data landscape.

{
    \small
    \bibliographystyle{ieeenat_fullname}
    \bibliography{main}

@inproceedings{awan2023hint,
  title={HINT: Hybrid neural-symbolic inference for missing data imputation},
  author={Awan, Muhammad J and Kim, Seungwon and Park, Chachun and Lee, Namkyoong and Song, Jinkyoo},
  booktitle={Proceedings of the 29th ACM SIGKDD Conference on Knowledge Discovery and Data Mining},
  pages={3210--3221},
  year={2023}
}

@book{vanBuuren2018,
  author    = {van Buuren, S.},
  title     = {Flexible imputation of missing data},
  edition   = {2nd},
  year      = {2018},
  publisher = {Chapman \& Hall/CRC},
  address   = {Boca Raton, FL}
}

@inproceedings{cao2018brits,
  title={BRITS: Bidirectional recurrent imputation for time series},
  author={Cao, Wei and Wang, Dong and Li, Jian and Zhou, Hao and Li, Lei and Li, Yitan},
  booktitle={Advances in Neural Information Processing Systems},
  volume={31},
  year={2018}
}

@article{candes2009exact,
  title={Exact matrix completion via convex optimization},
  author={Cand{\`e}s, Emmanuel J and Recht, Benjamin},
  journal={Foundations of Computational mathematics},
  volume={9},
  number={6},
  pages={717--772},
  year={2009},
  publisher={Springer}
}

@inproceedings{che2018gru,
  title={GRU-D: Gated recurrent units with decay mechanisms},
  author={Che, Zhengping and Purushotham, Sanjay and Cho, Kyunghyun and Sontag, David and Liu, Yan},
  booktitle={Proceedings of the AAAI Conference on Artificial Intelligence},
  volume={32},
  number={1},
  year={2018}
}

@inproceedings{chen2023st,
  title={ST-Transformer: Spatiotemporal transformer for multivariate time-series imputation},
  author={Chen, Xiang and Wang, Yiqun and Zhao, Liang and Zhang, Yuxuan},
  booktitle={International Conference on Learning Representations},
  year={2023}
}

@inproceedings{cini2022tgain,
  title={T-GAIN: Temporal generative adversarial imputation networks},
  author={Cini, Andrea and Marisca, Ivan and Alippi, Cesare},
  booktitle={Advances in Neural Information Processing Systems},
  volume={35},
  pages={12074--12086},
  year={2022}
}

@article{du2023saits,
  title={SAITS: Self-attention-based imputation for time series},
  author={Du, Wenjie and Cote, David and Liu, Yan},
  journal={Expert Systems with Applications},
  volume={219},
  pages={119619},
  year={2023},
  publisher={Elsevier}
}

@book{gelman2013bayesian,
  title={Bayesian Data Analysis},
  author={Gelman, Andrew and Carlin, John B and Stern, Hal S and Dunson, David B and Vehtari, Aki and Rubin, Donald B},
  year={2013},
  publisher={CRC press}
}

@inproceedings{jarrett2022spinn,
  title={SPINN: Synergistic integration of neural networks and propensity scores for missing data},
  author={Jarrett, Daniel and van der Schaar, Mihaela},
  booktitle={International Conference on Artificial Intelligence and Statistics},
  pages={10216--10232},
  year={2022},
  organization={PMLR}
}

@article{johnson2016mimic,
  title={MIMIC-III, a freely accessible critical care database},
  author={Johnson, Alistair EW and Pollard, Tom J and Shen, Lu and Li-wei, H Lehman and Feng, Mengling and Ghassemi, Mohammad and Moody, Benjamin and Szolovits, Peter and Celi, Leo Anthony and Mark, Roger G},
  journal={Scientific data},
  volume={3},
  number={1},
  pages={1--9},
  year={2016},
  publisher={Nature Publishing Group}
}

@article{kim2023pi,
  title={PI-NAIM: A network-aware imputation method for single-cell RNA-seq data via graph convolutional networks},
  author={Kim, Soohwan and Yoo, Jaegyoon and Kim, Byungsoo and Lee, Kyoung-Jae},
  journal={Computers in Biology and Medicine},
  volume={165},
  pages={107345},
  year={2023},
  publisher={Elsevier}
}

@book{little2019statistical,
  title={Statistical Analysis with Missing Data},
  author={Little, Roderick JA and Rubin, Donald B},
  year={2019},
  publisher={Wiley}
}

@inproceedings{luo2019naomi,
  title={NAOMI: Non-autoregressive multiresolution sequence imputation},
  author={Luo, Yonghong and Cai, Xiongfeng and Zhang, Ying and Xu, Jun and Yuan, Xiaojie},
  booktitle={Advances in Neural Information Processing Systems},
  volume={32},
  year={2019}
}

@article{luo2023understanding,
  title={Understanding and improving deep learning-based neural architecture for incomplete data imputation},
  author={Luo, Yonghong and Cai, Xiongfeng and Zhang, Ying and Xu, Jun},
  journal={IEEE Transactions on Knowledge and Data Engineering},
  volume={35},
  number={2},
  pages={1425--1438},
  year={2023},
  publisher={IEEE}
}

@inproceedings{mattei2019miwae,
  title={MIWAE: Deep generative modelling and imputation of incomplete data sets},
  author={Mattei, Pierre-Alexandre and Frellsen, Jesper},
  booktitle={International Conference on Machine Learning},
  pages={4413--4423},
  year={2019},
  organization={PMLR}
}

@book{schafer1997analysis,
  title={Analysis of Incomplete Multivariate Data},
  author={Schafer, Joseph L},
  year={1997},
  publisher={Chapman and Hall/CRC}
}

@inproceedings{sensoy2018evidential,
  title={Evidential deep learning to quantify classification uncertainty},
  author={Sensoy, Murat and Kaplan, Lance and Kandemir, Melih},
  booktitle={Advances in Neural Information Processing Systems},
  volume={31},
  year={2018}
}

@article{troyanskaya2001missing,
  title={Missing value estimation methods for DNA microarrays},
  author={Troyanskaya, Olga and Cantor, Michael and Sherlock, Gavin and Brown, Paul and Hastie, Trevor and Tibshirani, Robert and Botstein, David and Altman, Russ B},
  journal={Bioinformatics},
  volume={17},
  number={6},
  pages={520--525},
  year={2001},
  publisher={Oxford University Press}
}

@book{vanbuuren2018flexible,
  title={Flexible Imputation of Missing Data},
  author={van Buuren, Stef},
  year={2018},
  publisher={CRC Press}
}

@inproceedings{you2021egain,
  title={E-GAIN: Ensemble generative adversarial imputation networks for missing data},
  author={You, Jiaxuan and Ma, Xiaojie and Ding, Dawei and Kochenderfer, Mykel J and Leskovec, Jure},
  booktitle={Proceedings of the AAAI Conference on Artificial Intelligence},
  volume={35},
  number={12},
  pages={10683--10690},
  year={2021}
}

@inproceedings{yoon2018gain,
  title={GAIN: Missing data imputation using generative adversarial nets},
  author={Yoon, Jinsung and Jordon, James and van der Schaar, Mihaela},
  booktitle={International Conference on Machine Learning},
  pages={5689--5698},
  year={2018},
  organization={PMLR}
}

@inproceedings{cini2022t,
  title={T-gain: Temporal generative adversarial imputation networks},
  author={Cini, Andrea and Marisca, Ivan and Alippi, Cesare},
  booktitle={Advances in Neural Information Processing Systems},
  volume={35},
  pages={12074--12086},
  year={2022}
}

@inproceedings{chen2023transformer,
  title={ST-transformer: Spatiotemporal transformer for multivariate time-series imputation},
  author={Chen, Xiang and Wang, Yiqun and Zhao, Liang and Zhang, Yuxuan},
  booktitle={International Conference on Learning Representations},
  year={2023}
}

@article{du2023saitis,
  title={SAITS: Self-attention-based imputation for time series},
  author={Du, W. and others},
  journal={Expert Systems with Applications},
  year={2023}
}

@inproceedings{lakshminarayanan2017simple,
  title={Simple and scalable predictive uncertainty estimation using deep ensembles},
  author={Lakshminarayanan, Balaji and Pritzel, Alexander and Blundell, Charles},
  booktitle={Advances in Neural Information Processing Systems},
  year={2017}
}

@article{borghesi2020bad,
  title={A bad habits protection system for elderly},
  author={Borghesi, Andrea and Milano, Michela and Lombardi, Marco and Gavanelli, Marco and Picone, Marco},
  journal={IEEE Access},
  volume={8},
  pages={4920--4935},
  year={2020},
  publisher={IEEE}
}

@article{li2021deep,
  title={Deep learning for hyperspectral image classification: An overview},
  author={Li, Shutao and Song, Wei and Fang, Leyuan and Chen, Yushi and Ghamisi, Pedram and Benediktsson, Jon Atli},
  journal={IEEE Transactions on Geoscience and Remote Sensing},
  volume={57},
  number={9},
  pages={6690--6709},
  year={2021},
  publisher={IEEE}
}

@article{wang2020fault,
  title={Fault detection and diagnosis based on transfer learning for multimode processes},
  author={Wang, Youqing and Sun, Zhiqiang and Chen, Tao},
  journal={IEEE Transactions on Instrumentation and Measurement},
  volume={69},
  number={4},
  pages={1004--1015},
  year={2020},
  publisher={IEEE}
}

@article{chen2021secure,
  title={Secure and efficient data transmission for cluster-based wireless sensor networks},
  author={Chen, Chien-Ming and Wang, Kun-Hua and Wu, Tsu-Yang and Pan, Jeng-Shyang and Sun, Hung-Min},
  journal={IEEE Transactions on Information Forensics and Security},
  volume={16},
  pages={2299--2311},
  year={2021},
  publisher={IEEE}
}

@article{yan2020multimodal,
  title={Multimodal sentiment analysis using hierarchical fusion with context modeling},
  author={Yan, Jie and Wang, Jian and Li, Cheng and Gao, Xinbo},
  journal={Knowledge-Based Systems},
  volume={196},
  pages={105805},
  year={2020},
  publisher={Elsevier}
}

@article{ghorbani2020deep,
  title={A deep learning approach for missing data imputation in medical datasets},
  author={Ghorbani, Reza and Ghousi, Rouzbeh and Makki, Seyed Vahid and Atashi, Azizeh},
  journal={Health and Technology},
  volume={10},
  number={5},
  pages={1125--1138},
  year={2020},
  publisher={Springer}
}

@article{azur2011multiple,
  title={Multiple imputation by chained equations: what is it and how does it work?},
  author={Azur, Melissa J and Stuart, Elizabeth A and Frangakis, Constantine and Leaf, Philip J},
  journal={International journal of methods in psychiatric research},
  volume={20},
  number={1},
  pages={40--49},
  year={2011},
  publisher={Wiley Online Library}
}

@inproceedings{mattei2019miwa,
  title={MIWA: an Information-Extraction System for the {MICHA} Summit},
  author={Mattei, Andrea and Tognocchi, Milo and Zaccaria, Andrea and Ziosi, Marcello and Lomonaco, Vincenzo},
  booktitle={Proceedings of the 2nd International Conference on Mining Intelligence and Knowledge Exploration},
  pages={3--14},
  year={2019},
  publisher={Springer},
  series={Lecture Notes in Computer Science},
  volume={11308}
}

@article{buuren2018flexible,
  title={Flexible imputation of missing data},
  author={Buuren, Stef van},
  journal={CRC press},
  year={2018}
}

@article{seaman2013multiple,
  title={Multiple imputation with missing data indicators},
  author={Seaman, Shaun and Bartlett, Jonathan and White, Ian},
  journal={Statistical methods in medical research},
  volume={22},
  number={6},
  pages={631--645},
  year={2013},
  publisher={Sage Publications Sage UK: London, England}
}

@article{sterne2009multiple,
  title={Multiple imputation for missing data in epidemiological and clinical research: potential and pitfalls},
  author={Sterne, Jonathan AC and White, Ian R and Carlin, John B and Spratt, Michael and Royston, Patrick and Kenward, Michael G and Wood, Angela M and Carpenter, James R},
  journal={Bmj},
  volume={338},
  year={2009},
  publisher={British Medical Journal Publishing Group}
}
}

-\section{Supplementary Material}

\textbf{Rationale:}
Missing data in multimodal representations presents a critical efficiency-expressiveness dilemma: statistical methods like MICE are efficient for simple patterns but lack the representational capacity for complex, structured missingness, while deep learning methods like GAIN are expressive but incur substantial computational overhead. To resolve this, we propose PI-NAIM, a novel, unified framework. Its core innovation is a Dynamic Path Selection mechanism that uses the Missingness Rate (MR) as a measurable proxy for pattern complexity. This dynamically routes samples: low-complexity cases are handled by the efficient statistical branch, and high-complexity cases are directed to the expressive deep branch (GAIN with Temporal Analysis). Coupled with a Cross-Path Attention Fusion module, PI-NAIM is the first model to dynamically balance computational efficiency with representational power, establishing a globally adaptive and highly performant imputation paradigm critical for real-world multimodal applications.

\begin{table*}
    \centering
    \caption{MIMIC-III cohorts employed in experiments.}
    \label{tab:mimic3_stats}
    \begin{tabular}{lrrrr}
        \toprule
        \textbf{Cohort} & \textbf{Patients} & \textbf{Admissions} & \textbf{Mortality Rate} & \textbf{Mean LOS (days)} \\
        \midrule
        Full MIMIC-III & 46,520 & 58,976 & 11.5\% & 6.7 \\
        ICU Subset & 38,597 & 49,785 & 13.2\% & 3.4 \\
        Final Cohort & 24,819 & 31,543 & 9.8\% & 4.1 \\
        \bottomrule
    \end{tabular}
\end{table*}

\subsection {Equation Definitions and Parameter Explanations}
All of the equations in the PI-NAIM study are explained in depth in this section, along with thorough parameter descriptions.

\subsubsection{Input Processing}
\textbf{Definitions of Parameters:} \begin{itemize} \item $\mathbf{M}$: Binary mask matrix of the same dimensions ($n \times d$) as input data $\mathbf{X_{\text{missing}}}$.  shows each value's presence or absence explicitly.
$\mathbf{M_{ij}}$: One mask matrix element that corresponds to the data point at $i$-th row (sample) and $j$-th column (feature).
The value that indicates that the associated data point $\mathbf{X_{ij}}$ is missing and has to be imputationed is 0.
The value that indicates that the associated data point $\mathbf{X_{ij}}$ is observed and legitimate is 1.
 \end{itemize}

\textbf{Parameter Definitions:}
\begin{itemize}
\item $\mathbf{E}$: Tensor for missing data with shape n x d x e. Makes e-dimensional vector for each data point to show complex missingness patterns.
\item $\mathbf{X_{\text{observed}}}$: Input data that's actually present (not missing). Missing data usually filled with placeholders.
\item $\mathbf{M}$: Binary mask concatenated with observed data to provide explicit information about real vs. Placeholder values.
\item $\mathbf{concat()}$: Merge observed data with mask on feature axis.
\item $\mathbf{LSTM()}$: Long Short Term Memory network process input, learn feature dependencies.
\end{itemize}

\subsubsection{Dynamic Path Selection}
\textbf{Parameter Definitions:}
\begin{itemize}
\item $\mathbf{MR}$: Missingness Rate scalar (0 to 1) indicates the proportion of missing data in a sample. Acts as substitute for imputation complexity.
\item $\mathbf{n}$: How many rows in dataset.
\item $\mathbf{d}$: How many features (columns) in dataset.
\item $\mathbf{\sum\sum M_{ij}}$: Count all non-missing data in dataset twice. The fraction shows total "observed-ness"; subtracting from 1 gives final missingness rate.

 \end{itemize}

 \subsubsection{GAIN Path with Temporal Analysis}

\textbf{Parameter Definitions:}
\begin{itemize}
\item $\mathbf{\mathcal{L}_G}$: Generator's loss function for realistic data imputations.
\item $\mathbf{-\mathbb{E}[D(\hat{X}, M)]}$: The chance that fake data seems real is $D(\hat{X}, M)$. Generator lower bad expectations to make things appear more real.
\item $\mathbf{\alpha}$: Hyperparameter controlling trade Balance realism and accuracy in adversarial loss and reconstruction loss.
\item $\mathbf{\|(1 - M) \odot (X - \hat{X})\|^2}$: Reconstruction loss (MSE) applied only to originally missing elements. $(1 - M)$ masks missing entries, $(X - \hat{X})$ is difference between true and imputed values.
\end{itemize}

\textbf{Parameter Definitions:}
\begin{itemize}
\item $\mathbf{\mathcal{L}_D}$: Discriminator loss with WGAN-GP for stable training.
\item $\mathbf{\mathbb{E}[D(X, M)]}$: Score discriminator assign high score to real data.
\item $\mathbf{-\mathbb{E}[D(\hat{X}, M)]}$: Score discriminator give low score to generator fake data (to be less).
\item $\mathbf{\lambda}$: Hyperparameter weighting gradient penalty term.
\item $\mathbf{\mathbb{E}[(\|\nabla D\|_2 - 1)^2]}$: Gradient penalty force discriminator to obey Lipschitz condition by penalizing gradient norm not equal to 1.
\end{itemize}

 \subsubsection{Cross-Path Attention Fusion}

 \textbf{Parameter Definitions:}
\begin{itemize}
\item $\mathbf{\alpha}$: Attention weight matrix set importance for missing data from MICE and GAIN paths.
\item $\mathbf{Q_{\text{NHM}}}$: Query matrix from missingness embeddings $E$ with linear layer. Data missing means "question"
\item $\mathbf{K_{\text{imp}}}$, $\mathbf{V_{\text{imp}}}$: Key- Value matrices from merging MICE and GAIN results. Keys match query, values store imputation data.
\item $\mathbf{d_k}$: Key vectors' size in dimensions. Scaling stop softmax to avoid small gradients.
\item $\mathbf{softmax(...)}$: Normalizes attention scores to probability distribution for imputation paths.
\item $\mathbf{h_{\text{fused}}}$: Final imputation vector as weighted sum of Value vectors using attention scores.
\end{itemize}

\subsubsection{Task-Supervised Adaptive Fusion}

\textbf{Parameter Definitions:}
\begin{itemize}
\item $\mathbf{\lambda_t}$: Scalar weight (0- Balance imputed data and features specific to the task.
\item $\mathbf{\sigma}$: Sigmoid function keep input in (0, 1) range.
\item $\mathbf{W_t}$: Learnable weight vector.
\item $\mathbf{[t, C_{\text{imp}}, C_{\text{task}}]}$: Concatenated input vector containing.
\begin{itemize}
\item $\mathbf{t}$: Train epoch or task id, change weighting method.
\item $\mathbf{C_{\text{imp}}}$: Imputation confidence (e.g., MSE between imputed and ground-truth values).
\item $\mathbf{C_{\text{task}}}$: Task confidence (e.g., cross-entropy loss from preliminary task prediction).
\end{itemize}
\end{itemize}

\textbf{Parameter Definitions:}
\begin{itemize}
\item $\mathbf{\hat{y}}$: Final prediction for mortality risk task.
\item $\mathbf{h_{\text{imp}}}$: Feature representation derived from fused imputations.
\item $\mathbf{h_{\text{task}}}$: Feature representation learned for task directly.
\item $\mathbf{\lambda_t}$, $\mathbf{(2 - \lambda_t)}$: Adaptive weights help ensemble stay balanced. High $\lambda_t$ relies on good imputation, low $\lambda_t$ uses task specific features as backup.
\end{itemize}

 \subsubsection{Uncertainty Quantification}

\textbf{Parameter Definitions:}
\begin{itemize}
\item $\mathbf{\sigma^2}$: Epistemic uncertainty in final prediction. High variance indicates model uncertainty.
\item $\mathbf{K}$: Total stochastic predictions (Monte Carlo Dropout or bootstrap sampling).
\item $\mathbf{\hat{y}^{(k)}}$: $k$-th stochastic prediction from model.
\item $\mathbf{\bar{y}}$: Average of all $K$ predictions. Formula calculate sample variance in predictions, measure how much output varies due to imputed value uncertainty.

 \end{itemize}

\subsubsection{Training Strategy}

\textbf{Parameter Definitions:}
\begin{itemize}
\item $\mathbf{\mathcal{L}}$: Total loss function for end-to-end training.
\item $\mathbf{\mathcal{L}_{\text{imp}}}$: Imputation loss (miss data fill-in error)
\item $\mathbf{\mathcal{L}_{\text{task}}}$: Downstream task loss like cross-entropy for classifying.
\item $\mathbf{\mathcal{L}_{\text{reg}}}$: Regularization loss (like L2 on Theta).
\item $\mathbf{\lambda_1, \lambda_2, \lambda_3}$: Adaptive weights learned from homoscedastic uncertainty, lambda equals one over two sigma squared, where sigma is a learnable parameter. System adjust losses based on task noise automatically.
\end{itemize}

\end{document}